\documentclass[letterpaper, 10 pt, conference]{ieeeconf}
\usepackage{booktabs}
\IEEEoverridecommandlockouts                              % This command is only needed if 
\usepackage{graphics} % for pdf, bitmapped graphics files
\usepackage{epsfig} % for postscript graphics files
\usepackage{mathptmx} % assumes new font selection scheme installed
\usepackage{times} % assumes new font selection scheme installed
\usepackage{amsmath} % assumes amsmath package installed
\usepackage{amssymb}  % assumes amsmath package installed
\usepackage[ruled,vlined]{algorithm2e}
\usepackage{hyperref}
\usepackage[table,xcdraw]{xcolor}
\usepackage{cite}

\title{\LARGE \bf
Residual Reactive Navigation: Combining Classical and Learned Navigation Strategies For Deployment in Unknown Environments}

\author{Krishan Rana, Ben Talbot, Vibhavari Dasagi, Michael Milford, and Niko S\"{u}nderhauf% <-this % stops a space
\thanks{This research was partially supported by funding from ARC grants FT140101229, CE140100016 and the QUT Centre for Robotics.}% <-this % stops a space
\thanks{The authors are with the QUT Centre for Robotics, School of Electrical Engineering and Robotics and the Australian Centre for Robotic Vision at the Queensland University of Technology.
        {\tt\small krishan.rana@hdr.qut.edu.au}}%
}

\begin{document}

\maketitle
\thispagestyle{empty}
\pagestyle{empty}

%%%%%%%%%%%%%%%%%%%%%%%%%%%%%%%%%%%%%%%%%%%%%%%%%%%%%%%%%%%%%%%%%%%%%%%%%%%%%%%%

\begin{abstract}

In this work we focus on improving the efficiency and generalisation of learned navigation strategies when transferred from its training environment to previously unseen ones. We present an extension of the residual reinforcement learning framework from the robotic manipulation literature and adapt it to the vast and unstructured environments that mobile robots can operate in. The concept is based on learning a residual control effect to add to a typical sub-optimal classical controller in order to close the performance gap, whilst guiding the exploration process during training for improved data efficiency. We exploit this tight coupling and propose a novel deployment strategy, \textit{switching} Residual Reactive Navigation (sRRN), which yields efficient trajectories whilst probabilistically switching to a classical controller in cases of high policy uncertainty. Our approach achieves improved performance over end-to-end alternatives and can be incorporated as part of a complete navigation stack for cluttered indoor navigation tasks in the real world. The code and training environment for this project is made publicly available at \url{https://sites.google.com/view/srrn/home}.

\end{abstract}

%%%%%%%%%%%%%%%%%%%%%%%%%%%%%%%%%%%%%%%%%%%%%%%%%%%%%%%%%%%%%%%%%%%%%%%%%%%%%%%%
\section{Introduction}

Deep reinforcement learning approaches have been shown to learn efficient and complex reactive navigation policies, portraying behaviours which are difficult to derive analytically. They however require extensive amounts of online training data which is a limiting factor for real robot applications. Additionally, at deployment, these systems could yield undesirable behaviour in states where the policy failed to generalise, particularly in novel, unseen environments. This reduces the overall reliability we can place on these systems. On the other hand, classical approaches in reactive robotic navigation, are generally deterministic and can portray competent obstacle avoidance capabilities in any given domain. However, lack the complexity required to efficiently navigate in cluttered environments and are commonly plagued by oscillations \cite{khatib1986real}, seizure in local minima \cite{koren1991potential} and poor path efficiency. Such high level solutions are difficult to hand-engineer explicitly and tend to deteriorate in performance when extensively tuned for a particular environment.

\begin{figure}[thpb]
  \centering
  \includegraphics[width=0.49\textwidth]{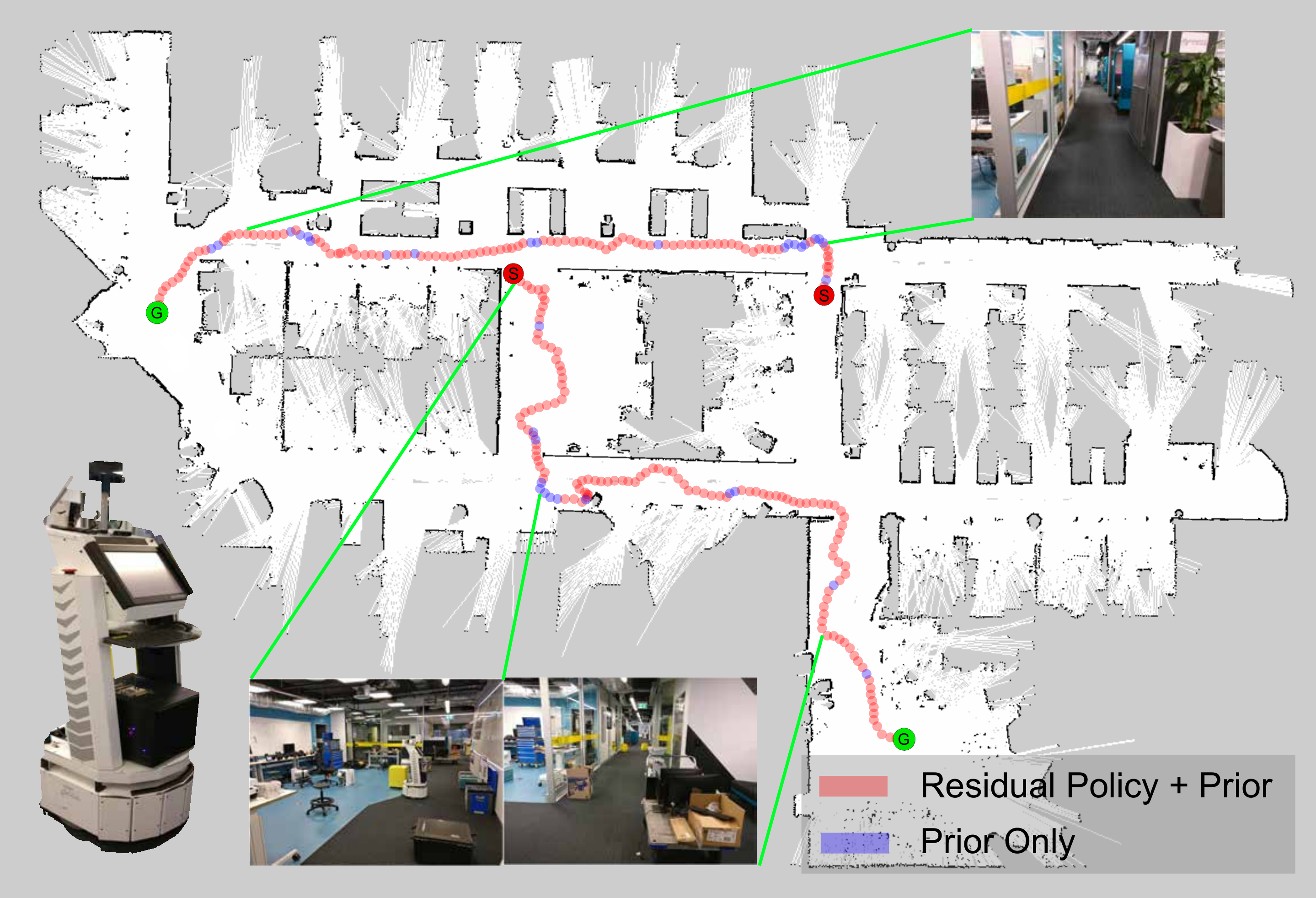}
  \caption{Successful trajectory achieved by sRRN in a real world cluttered indoor environment running as the local component of the ROS navigation stack. We colour code the trajectory to indicate which control effect was executed in the environment. Purple regions indicate execution of prior only in events of high policy uncertainty. The red regions indicate execution of the combined systems for improved navigation performance.}
  \label{front}
  \vspace{-0.3cm}
\end{figure}

In this work we attempt to combine the classical and learned strategies in order to address their respective limitations. We build upon prior work in the area of residual reinforcement learning, which primarily focuses on robotic manipulation tasks, and adapt it to the goal-directed navigation task presented by Anderson \textit{et al.} \cite{anderson2018evaluation}. The concept is based on learning to improve upon prior experience rather than learning the entire task structure from scratch. We train a continuous drive mobile robot equipped with a 180$^{\circ}$ laser scanner to efficiently navigate towards arbitrary goals in cluttered environments by learning a residual policy to close the performance gap of a classical reactive controller (prior). All training is completed in simulation and zero shot transferred to a real robot. We present a novel deployment strategy that exploits this tight coupling of the two systems and probabilistically falls back to the classical controller in cases of high policy uncertainty as shown in Figure \ref{front}. This reduces the chances of undesirable and potentially catastrophic failure. We show a significant improvement to the variance and training sample efficiency required to learn the navigation task and superior generalisation performance at deployment on a real robot when compared to end-to-end learned systems, the baseline prior and the commonly used move-base ROS navigation stack. We created a lightweight simulation environment for this navigation task and release it together with our code for this work at \url{https://sites.google.com/view/srrn/home}.\\
To summarise, the key contributions of this paper are:
\begin{itemize}
    \item a novel deployment strategy for continuous control reactive navigation agents which incorporates model uncertainty to stochastically inform a switching condition when the policy is uncertain.
    \item an approach which tightly couples classical reactive navigation strategies with learning-based methods, showing significant reductions to the sample complexity and variance when learning to navigate end-to-end.
    \item we show that our approach can directly be transferred from simulation to the real-world for cluttered indoor navigation tasks without any fine-tuning and can outperform end-to-end trained systems.
    
\end{itemize}

\section{Related Work}
% Vibha: Why is it "reactive"? RL is quite active.
\subsection{Learning for Reactive Navigation}
With the recent surge in deep learning, several supervised and reinforcement learning based approaches have been presented for the task of reactive robot navigation. Supervised approaches for obstacle avoidance have been shown to enable a robot to navigate along forest trails using an imitation learning approach \cite{ross2013learning}. Kim \textit{et al.} \cite{kim2015deep} and Pfeiffer \textit{et al.} \cite{pfeiffer2017perception} train a navigation agent using a global path planner as labelled training data. These systems are however limited to the performance of the training data. Zhu \textit{et al.} \cite{zhu2017target} train a monocular-based robot, using deep reinforcement learning for target driven navigation. Given the significant amount of environment interaction required, the agent is trained in a high fidelity simulator and then fine tuned before deployment in the real world. Given the close correspondence of laser scans in simulation and the real-world, \cite{tai2017virtual} and \cite{xie2018learning}  show that training an agent in simulation can be transferred directly to a real robot. % Vibha: Don't make this next part sound so negative. You want to acknowledge the capabilities of previous works while gently talking about the limitations.
Despite showing reasonable performance, the robot was still shown to fail in scenarios it did not generalise well to. Additionally both approaches utilise a dense reward signal to accelerate the training process which we found to yield inefficient routes to the target.

\subsection{Classical Navigation Approaches}
Navigation is a well developed field in the robotics community and can be divided into two categories: \textit{deliberative} and \textit{reactive} systems. While deliberative systems typically rely on the availability of a globally consistent map, % The deliberative systems rely on heavy reasoning and detailed representation for decision making. Such system are typically approached using Simultaneous Localisation and Mapping (SLAM) \cite{thrun2005probabilistic} \cite{bailey2006simultaneous} to obtain a globally consistent geometric map of the environment within which a path is found. 
reactive planners determine collision free paths based on immediate perception of their surrounding environment. They can handle dynamic obstacles and those unaccounted for in the global map. Potential field approaches \cite{warren1989global}\cite{hwang1992potential} are a popular set of reactive algorithms which utilise a local potential function to attract the robot towards the goal and repel it away from obstacles. Vector Field Histogram (VHF)~\cite{borenstein1991vector} is a real time motion planning algorithm that generates a polar histogram to represent the polar density of obstacles around the robot. The robot's steering angle is then chosen based on least polar density and closeness to goal. Such an approach requires the the polar histogram to be computed at every timestep and is suited for sparse moving obstacles. Other approaches utilise short term memory to inherently build a local map \cite{antich2005extending}, allowing them to escape from local minima. The \textit{bug} family of algorithms \cite{ng2007performance} can guarantee completeness but is inefficient. A common disadvantage of these classical approaches is that they require significant tuning and hand engineering to achieve good performance, coupled with a tendency to deteriorate when deployed in domains with different characteristics. Further common problems are the occurrence of local minima \cite{koren1991potential}, oscillations \cite{khatib1986real} and suboptimal path length.

% All these approaches require significant tuning and hand engineering in order for them to work well in a given environment and tend to deteriorate in performance across domains when extensively tuned for a particular environment \cite{koren1991potential}. Additionally most of these approaches face common problems related to seizure in local minimum \cite{koren1991potential}, oscillations \cite{khatib1986real} and poor path efficiency. Such solutions are difficult to derive analytically.

\subsection{Combining Classical and Learned Systems}
There has been a recent interest in the way of combining the best of both worlds in the robotics community: classical and learned systems. Bansal \textit{et al.} \cite{bansal2019combining} train a perception module to produce obstacle-free waypoints with which an optimal controller can follow. Xie \textit{et al.} \cite{xie2018learning} leverage a simple classical controller in form of a P-controller to speed up the training process for a navigation agent. The idea is based on the fact that a simple P-controller will yield higher rewards than random exploration alone. More closely related to this work is the residual reinforcement learning approach introduced by Johannink \textit{et al.} \cite{johannink2019residual} and Silver \textit{et al.} \cite{silver2018residual} concurrently. The approach is based on learning a residual policy to improve upon a sub-optimal classical controller. Both works focused on robotic manipulation tasks where the residual learned to deal with contacts and friction, typically difficult to model analytically. We extend the approach for the first time % Vibha: to our knowledge
to the navigation domain, which exhibits a larger unstructured operational space. We particularly focus on the deployment of these system where we take advantage of the close coupling of the learned and classical controller as a result of the joint training process. %Other works in this area. Vibha: more references here 
Other works in this area \cite{Iscen2018PoliciesMT} show how a learned policy can be used to modulate Trajectory Generators in order to improve upon their base level of performance and show its applicability for real robot locomotion. In all cases we see how the learned system benefit from prior knowledge during training and the prior systems benefit from the learned system during deployment. However there has been little work in leveraging this close coupling to improve the generalisation capabilities of these system when deployed in novel environments. In this work we take a step towards this notion.

%Vibha: end with a sentence or two reiterating gaps and how you intend to address them

\section{Background}
In this work we attempt to exploit the efficiency and adept capabilities provided by learned strategies in reactive navigation whilst improving the generalisation of these systems when transferred to the real world. We closely couple the learned policy with a classical controller using residual reinforcement learning and utilise an uncertainty measure to inform a stochastic switching condition in cases of high policy uncertainty. We introduce the navigation task and key components of our strategy in this section before describing our approach in detail.

% \begin{figure}[thpb]
%   \centering
%   \includegraphics[width=0.5\textwidth]{Images/path_evaluation_1.png}
%   \caption{Path Evaluation}
%   \label{path1}
% \end{figure}

\subsection{Problem Formulation}
We formulate the problem as a decision making process where an agent is required to avoid obstacles and reach a target location in the shortest time possible in unknown environments. Given an input $x_{t}$ at time $t \in [0,T]$, the agent is required to determine a suitable action $a_{t} \in A$. After executing the action, the robot transitions to the next state $x_{t+1}$ and receives a reward $r_{t}$ from the environment. We found the standard dense reward used in prior work \cite{tai2017virtual,xie2018learning,xie2017towards} to bias the policy towards learning long sub-optimal paths to the goal in its attempt to maximise its reward. Given our motivation to learn an efficient planner, we focus on the sparse reward setting, whilst being significantly difficult, allows the policy to identify novel and potentially efficient solutions. 
We define the reward as $r(s_{t}, a_{t}) = 1$ if $d_{target} < d_{threshold}$ and $r(s_{t}, a_{t}) = 0$ otherwise,
% We define the reward signal as follows:
% \[
%     r(s_{t}, a_{t})= 
% \begin{cases}
%     1 ,& \text{if } d_{target} < d_{threshold}\\
%     0 ,& \text{otherwise}
% \end{cases}
% \]
where $d_{target}$ is the distance between the target and the agent and $d_{threshold}$ is a set threshold. The overall objective of the decision making process is to maximise the cumulative future rewards $\sum^{T}_{\tau=t}\gamma^{\tau-1}r_{\tau}$, where $\gamma$ is a discount factor.
%Vibha: I'd call it d_{target} and d_{threshold}

% is a 20 dimensional vector consisting of 15 binned laser scans, the priors linear and angular velocity, the previous actions executed in the environment and the distance and angle to the goal. The laser scan binning follows from Tai \textit{et .al} [X] who show that averaging the laser scans into these bins help deal with the noise associated with theses sensors and allow for reasonable transfer to a real robot. 

% \subsection{Twin Delayed Deep Deterministic Policy Gradients (TD3)}
% TD3 is an off-policy deep reinforcement learning algorithm which is part of the actor-critic family of algorithms. 

%Vibha: you don't need to mention by name. This applies to the related work section as well. Especially if you're trying to point out limitations it can come across as targeting them directly instead of using them as examples to make your point.
\subsection{Estimating Policy State Uncertainty}
There have been several approaches introduced in the deep learning literature which attempt to extract uncertainty from neural networks \cite{gal2016dropout,kleiner2012big,pearce2018uncertainty,lakshminarayanan2017simple}. With the extension of deep neural networks in the reinforcement learning community, recent work has explored the applicability of these approaches to deep reinforcement learning algorithms and has shown impressive results particularly in the area of safe reinforcement learning. Kahn \textit{et al.} \cite{kahn2017uncertainty} utilise a combination of MC-dropout \cite{gal2016dropout} and bootstrapping \cite{kleiner2012big,osband2016deep} to predict the epistemic uncertainty in a collision prediction model which modulates the agent's velocity. L\"{u}tjens \textit{et al.} \cite{Ltjens2018SafeRL} utilise a similar uncertainty estimation method to enable safe pedestrian avoidance. Clements \textit{et al.} \cite{clements2019estimating} introduce an approach which can separately determine the aleatoric risk using a distributional approach and epistemic uncertainty using a Bayesian framework. Other works have suggested the use of discriminative models to provide estimates of uncertainty \cite{daftry2016introspective}. Such estimates yield poor results in novel states which the model has not generalised well to, limiting its applicability.

\section{Residual Reactive Navigation}
%Vibha: This is where you're introducing your method for the first time. Try to make the distinction between what the method is and what its implementation is. Your method shouldn't be tied in to the RL algorithm, VFH, bin size, network size, environment size, etc. It's meant to be a general description of how to solve the problem. The implementation is an instance of your method. It's what you used to evaluate the efficacy of your method.
We extend the field of residual reinforcement learning to the navigation domain and exploit the tight coupling between a classical and learned controller to address the generalisation problem faced by trained systems when transferred to a different domain. We achieve this by utilising an uncertainty estimate of the residual policy to inform a switching condition to the prior in cases of high uncertainty. We firstly describe the training process of our approach and then describe our novel deployment strategy.

\subsection{Training}
\label{training}
We follow a similar process to Johannink \textit{et al.} \cite{johannink2019residual} for residual reinforcement learning. The system is decomposed into two components: a residual policy and a classical controller referred to as the \textit{prior}. 

The resulting action effect from the two systems can be described as $a_{t} = \pi_{\theta}(s_{t}) + \pi_{prior}(s_{t})$, where
% \[a_{t} = \pi_{\theta}(s_{t}) + \pi_{prior}(s_{t})\]
$\pi_{\theta}$ represents the learned residual policy and $\pi_{prior}$ represents the prior. We refer to this combination as a \textit{hybrid} action. The residual policy is parameterised by a neural network and produces a suitable action effect to modify the priors' output. For the prior, we utilise a variant of the Artificial Potential Fields controller introduced by Warren \textit{et al.} \cite{warren1989global}. It demonstrates a competent level of obstacle avoidance capabilities and exhibits the same limitations faced by most reactive planners \cite{borenstein1991vector}. We note here that any other prior controller for this task can be utilised. The policy is trained using TD3 \cite{fujimoto2018addressing}, an off-policy deep reinforcement learning algorithm for continuous action spaces in conjunction with the prior as described in \cite{johannink2019residual}. The output of the prior is a 2 dimensional vector consisting of a continuous linear, $v \in [0,1]$ $m/s$ and angular, $\omega \in [-1,1]$ $rad/s$ velocity. The output of the policy is a 2 dimensional vector consisting of a learned residual term for both the linear, $\delta v \in [-1,1]$ $m/s$  and angular, $\delta \omega \in [-1,1]$ $rad/s$ velocity. The range of the residual action space allows the network to negate the priors actions in states it deems necessary. To enable the extraction of the policies uncertainty at deployment, we train the actor network with \textit{MC-dropout} with a dropout ratio of 0.2. We note here that any epistemic uncertainty measure can be used here and the focus of this work is primarily on how it can be used within the given system. 
\subsubsection*{System Inputs}
We assume that the robot can localise itself within a global map in order to determine its relative position to a goal location. We divide the 180$^{\circ}$ laser scan range data into 15 bins and concatenate to it: the robots angle and distance to goal; previous linear and angular velocity; and the linear and angular velocities determined by the prior controller for the current state $s_{t}$. We found that this was a necessary step to improve the stability during training, which yielded smoother trajectories during deployment. The observation space is hence a 21 dimensional vector.

\subsection{Deployment}

Directly transferring a policy trained in simulation to the real world is typically plagued by the failure of the policy to generalise to unseen states. We propose a novel deployment strategy which builds on the RRN approach described above. The prior within the RRN system is enhanced by the learned residual; however in cases where the residual fails to generalise it could potentially deteriorate the performance of the system, limiting its reliability. We address this by introducing switching-RRN (sRRN) which probabilistically switches off the effect from the learned residual in cases of high policy uncertainty, instead only executing the prior. The stochastic switching is reminiscent of the $\epsilon$-greedy strategy used in multi-armed bandit problems \cite{katehakis1987multi} and removes the need for fixed thresholds which are difficult to determine beforehand. The higher the uncertainty, the higher the probability of choosing the \textit{prior} over the \textit{hybrid} action. The prior is a reliable fallback, which despite being sub-optimal, is deterministic and can be guaranteed to no have no collisions. The complete sRRN algorithm is given in Algorithm \ref{algorithm1}. %Vibha: reference the algorithm! Don't say "below".

% stochaisticity prevent our system from exhinibiting the same performance of the prior

% Vibha: use proper indentation for your if-else statement
\begin{algorithm}[t]
\label{algorithm1}
\SetAlgoLined
\textbf{Given:} Residual Policy ($\pi_{\theta}$), Prior Controller ($\pi_{prior}$) \\
\KwIn{State $s_{t}$}
\KwOut{Linear Velocity \textit{\textbf{v}}, Angular Velocity $\omega$}
  Compute actions using prior:\\
  $\textit{\textbf{v}}_{prior}$, $\omega_{prior}$\ = $\pi_{prior}(s_{t})$;\\
  Compute the mean and variance of residual actions using MC-Dropout:\\
  $\mu_{\delta v_{policy}}$, $\mu_{\delta \omega_{policy}} = \mathop{\mathbb{E}}[\pi_{\theta}(s_{t})]$\\
  $\sigma^{2}_{\delta v_{policy}}$, $\sigma^{2}_{\delta \omega_{policy}} = \mathrm{Var}[\pi_{\theta}(s_{t})]$\\
  With probability $\epsilon = max(\sigma^{2}_{\delta v_{policy}}, \sigma^{2}_{\delta \omega_{policy}})$, select prior action:\\
  \hspace{0.5cm}\textit{\textbf{v}} $= \textit{\textbf{v}}_{prior}$; $\omega = \omega_{prior}$ \\
%   \hspace{0.5cm}$\omega = \omega_{prior}$\\
  Otherwise:\\ 
  \hspace{0.5cm}\textit{\textbf{v}} $= \textit{\textbf{v}}_{prior} + \mu_{\delta v_{policy}}$; $\omega = \omega_{prior} + \mu_{\delta \omega_{policy}}$\\
%   \hspace{0.5cm}$\omega = \omega_{prior} + \mu_{\delta \omega_{policy}}$\\
 \Return {\textit{\textbf{v, $\omega$}}}
 \caption{sRRN}
\end{algorithm}

The residual policy action and corresponding uncertainty are extracted by finding the mean and variance of 100 stochastic forward passes of the actor network using \textit{MC-dropout}.

\section{Experiments}
\label{experiments}
% Vibha: first statement is redundant. You've already used the term RRN before so no need for that sentence.
We conducted several experiments to evaluate the performance of our approach in both simulation and the real world.  Due to the lack of an efficient environment dedicated to learning navigation strategies using 2D laser scanners, we created a custom environment using Box2D, a 2-dimensional physics simulation engine which provides a significant performance boost to the learning process over Gazebo and VREP used in prior work \cite{learn_nav} \cite{tai2017virtual}. We freely release our environment\footnote{\url{https://sites.google.com/view/srrn/home}} to enable further development in this area. Given the close correspondence of 2D laser scans in simulation and the real world, we perform zero shot transfer of our trained policy to a real robot and evaluate the effectiveness of the system when incorporated into the ROS navigation stack for cluttered indoor navigation. Our approach, which we term Residual Reactive Navigation (RRN) is divided in two variants at deployment: \textit{without switching} and \textit{with switching} (sRRN). The first approach executes the combined system at all stages during deployment and the second follows the switching strategy we introduce in Algorithm 1. We clip the resulting action outputs within the range [-1,1]. RRN is compared across 4 different baselines:\\
\textbf{End-to-end:} Using TD3, a policy is trained end-to-end to learn the entire navigation task from scratch. To keep consistent with reward definitions in this study we utilise the sparse reward signal for this method. The input to the network is a 19 dimensional state vector similar to that described in Section \ref{training}, except that it does not include the prior controller actions. Both actions are clipped to the range -1 and 1 by a \textit{tanh()} activation function.\\
\textbf{Prior:} This represents the analytically derived local navigation controller based on the Artificial Potential Fields approach \cite{choset2005principles}. The input to this system is the angle to the goal and the raw 180$^\circ$ laser scan data.\\ The priors linear velocity is limited to be between the range [0,1] whilst, its angular velocity spans the range [-1,1]. 
\textbf{Random: } Actions are randomly sampled from a uniform distribution from the range [-1,1].\\
% Vibha: If you're talking about -1 to 1 here, also mention the action ranges for the residual and hybrid actions.
% \textbf{CORE-RL \cite{cheng2019control}:} An approach reminiscent of the residual learning method however by slowly annealing the prior during the learning process, a single policy is learned which is independent of the prior. The input to the network is the 21 dimensional state vector described in Section [X].\\
\textbf{ROS Move-Base:} For the real robot experiments, we additionally provide a comparison to the default DWA planner in ROS move-base.\\

To evaluate the performance of these systems, we report the average episodic success which is achieved when the agent arrives within 0.2m of the goal. The episode is timed out after 300 timesteps and is considered unsuccessful thereafter. In addition to success rate, we report Success weighted by (normalised inverse) Path Length (SPL), a specific measure for the \textit{PointGoal} task presented by Anderson \textit{et al.} \cite{anderson2018evaluation}, and the episodic actuation time as a measure of efficiency of the approach. The SPL ratio requires a measure of the shortest path to the goal which we approximate using the path found by an A-Star search across a 2000 $\times$ 1000 grid.

\subsection{Evaluation of Training Performance}
\begin{figure}[t]
  \centering
  \includegraphics[width=0.5\textwidth]{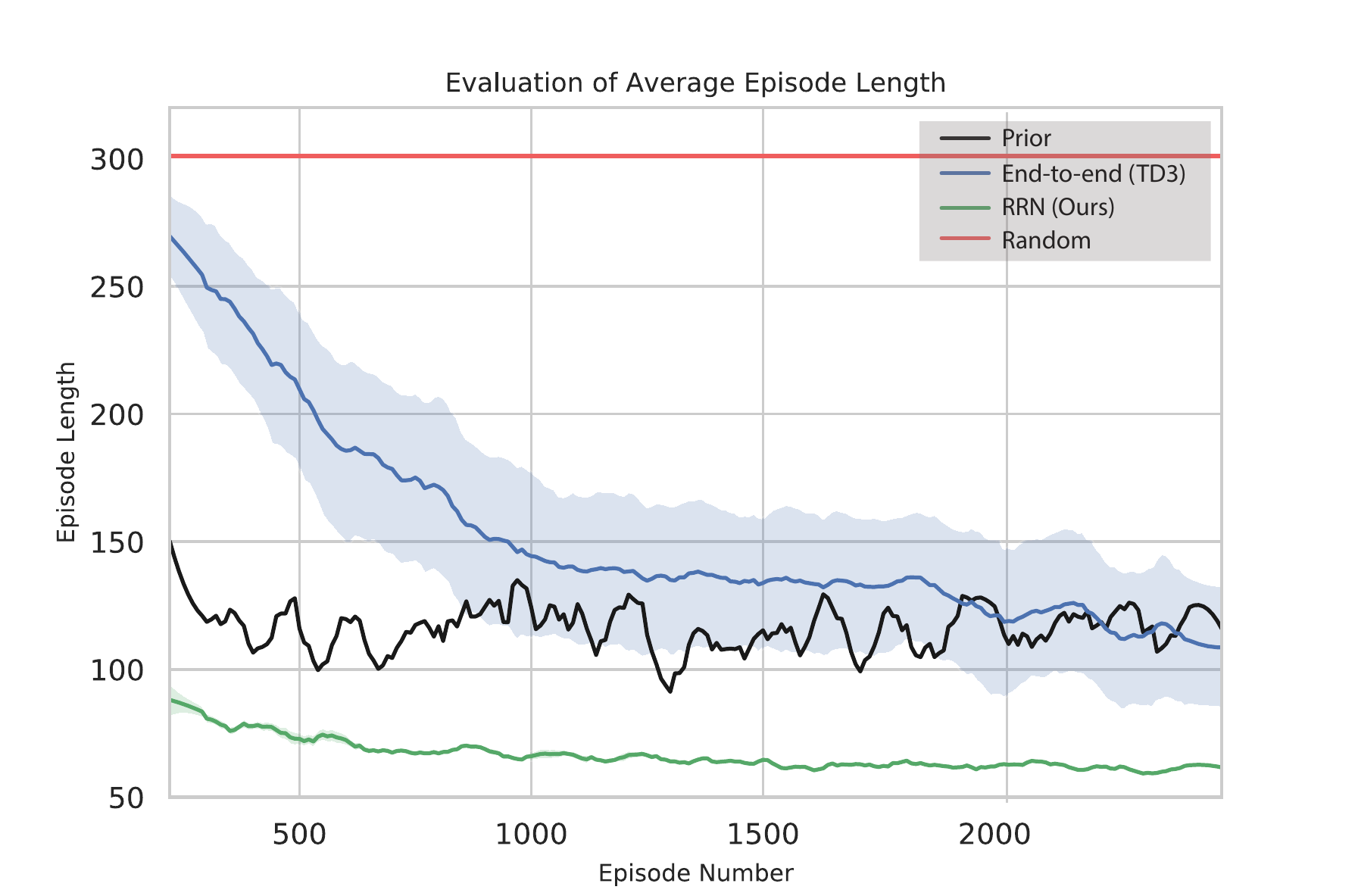}
  \caption{Average path length per episode during training}
  \label{avglength}
  \vspace{-0.5cm}
\end{figure}

We evaluated the training performance of each approach after every 10 episodes during training and the results are shown in Figure \ref{avglength}. All evaluations were made across 10 different seeds. The performance of the prior and random agent are overlayed for comparison. The end-to-end based approach shows considerably higher variance when compared to our approach, and requires a larger number of training episodes before convergence. The prior alone shows sub-optimal performance, but it is significantly better than a totally random agent which was incapable of achieving any reward. The residual learning approach shows significantly faster convergence which can be attributed to the prior guiding the exploration process, whilst allowing the policy to explore the regions around it for potential improved returns. In all the learning cases, the agent is driven by the sparse reward signal to identify the fastest route to the goal.

\subsection{Evaluation of Deployed Systems}
The main contribution in this work focuses on a novel stochastic deployment strategy which takes advantage of this close coupling to enable robust deployment in unseen environments. The evaluation is made both in simulation and on a real robot for target driven navigation tasks in cluttered indoor environments.

\subsubsection{Simulation Environment Evaluation}
We evaluate the agent in two different scenarios for its goal and domain generalisation capabilities. The goal generalisation is reported for the agent's performance to  navigate towards a set of unseen goal locations during training. This is done within the environments used for training. The environment generalisation performance is evaluated within a set of novel environments previously unseen to the agent. We present the results in Table \ref{tab:eval_table}.

\begin{table}[bt]
\centering
\caption{Evaluation within simulation environment}
\label{tab:eval_table}
\resizebox{0.49\textwidth}{!}{%
\begin{tabular}{@{}lccc
>{\columncolor[HTML]{C0C0C0}}c 
>{\columncolor[HTML]{C0C0C0}}c 
>{\columncolor[HTML]{C0C0C0}}c l@{}}
\toprule
 & \multicolumn{3}{c}{\textbf{Goal Generalisation}} & \multicolumn{3}{c}{\cellcolor[HTML]{C0C0C0}\textbf{Environment Generalisation}} &  \\ \midrule
\multicolumn{1}{c}{\textbf{Method}} & \textbf{\begin{tabular}[c]{@{}c@{}}Success \\ Rate\end{tabular}} & \textbf{SPL} & \textbf{\begin{tabular}[c]{@{}c@{}}Actuation\\ Time\end{tabular}} & \textbf{\begin{tabular}[c]{@{}c@{}}Success\\  Rate\end{tabular}} & \textbf{SPL} & \textbf{\begin{tabular}[c]{@{}c@{}}Actuation\\ Time\end{tabular}} &  \\ \midrule
Prior & 0.933 & 0.932 & 95.5 & 0.800 & 0.800 & 94.9 &  \\
End-to-end & 1.00 & 0.866 & 57.4 & 0.867 & 0.681 & 76.4 &  \\
\textbf{sRRN} & \textbf{1.00} & \textbf{0.954} & \textbf{68.2} & \textbf{0.933} & \textbf{0.920} & \textbf{73.9} &  \\
\textbf{RRN (without switching)} & \textbf{1.00} & \textbf{0.906} & \textbf{64.1} & \textbf{0.867} & \textbf{0.887} & \textbf{62.1} & \multicolumn{1}{c}{} \\
%CORE-RL & 0.933 & 0.895 & 55.5 & 0.733 & 0.651 & 71.1 &  \\
Random & 0.0667 & 0.148 & 278 & 0.00 & 0.00 & 300 &  \\ \bottomrule
\end{tabular}%
}
\vspace{-0.3cm}
\end{table}

\begin{figure}[t]
  \centering
  \includegraphics[width=0.49\textwidth]{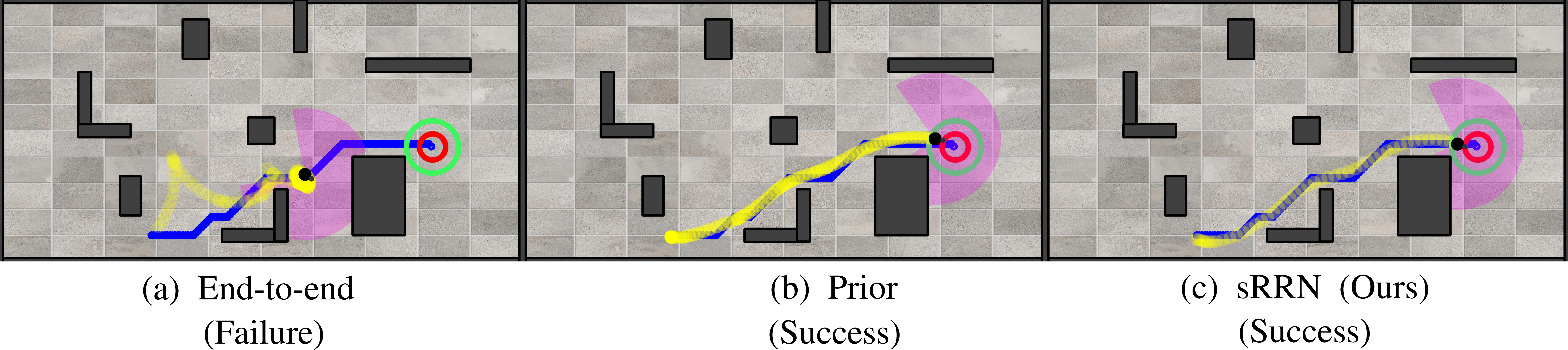}
  \caption{Evaluation of agent path in novel unseen environment. The dark blue line illustrates the shortest path identified by the A* Star planner and the yellow path illustrates the actual trajectory taken by the robot.}
  \label{traj}
  \vspace{-0.3cm}
\end{figure}

Our approach shows improved performance in all cases, achieving the highest success rate across both scenarios. We particularly note that the switching variant of RRN, as presented in Algorithm \ref{algorithm1}, plays a significant role when deploying the agent in novel unseen environments. This is attributed to the ability of the hybrid system (residual policy + prior) to fall back on to solely the prior in cases of high residual policy uncertainty. This is a limitation faced by the end-to-end approach which totally relies on the policy generalising to all unseen cases, which is not the case given finite training time in non-exhaustive environments. The prior, whilst showing significantly longer actuation times, is still capable of reaching the goal reliably as indicated by its high success rate. The high actuation time is a result of the oscillations the prior controller is prone to when navigating. Whilst there are possible analytical solutions that can solve this problem, we focus on showing how learning a residual can overcome this need and potentially learn far superior solutions. The high success rate of the prior makes it a dependable fall back for the sRRN system at the expense of an efficiency drop. This is a reasonable trade-off which improves the overall robustness of the system in novel environments. Figure \ref{traj} shows the trajectory taken by the 3 different systems.

 The trajectory (given in yellow) is plotted using transparency to illustrate the amount of time the agent took in a given location before progressing. We show a typical failure case experienced by end-to-end trained systems, which in the presence of an unfamiliar state, acts in an undesirable manner. Both the prior and our approach successfully reach the goal, however as shown by the more pronounced trajectory of the prior, the prior spent a longer period in a given location as a result of oscillations before progressing forward. This shows how the residual policy in RRN was capable of learning the appropriate control effect to modify the prior with in order to overcome such a drawback.
 
\begin{figure}[t]
  \centering
  \includegraphics[width=0.49\textwidth]{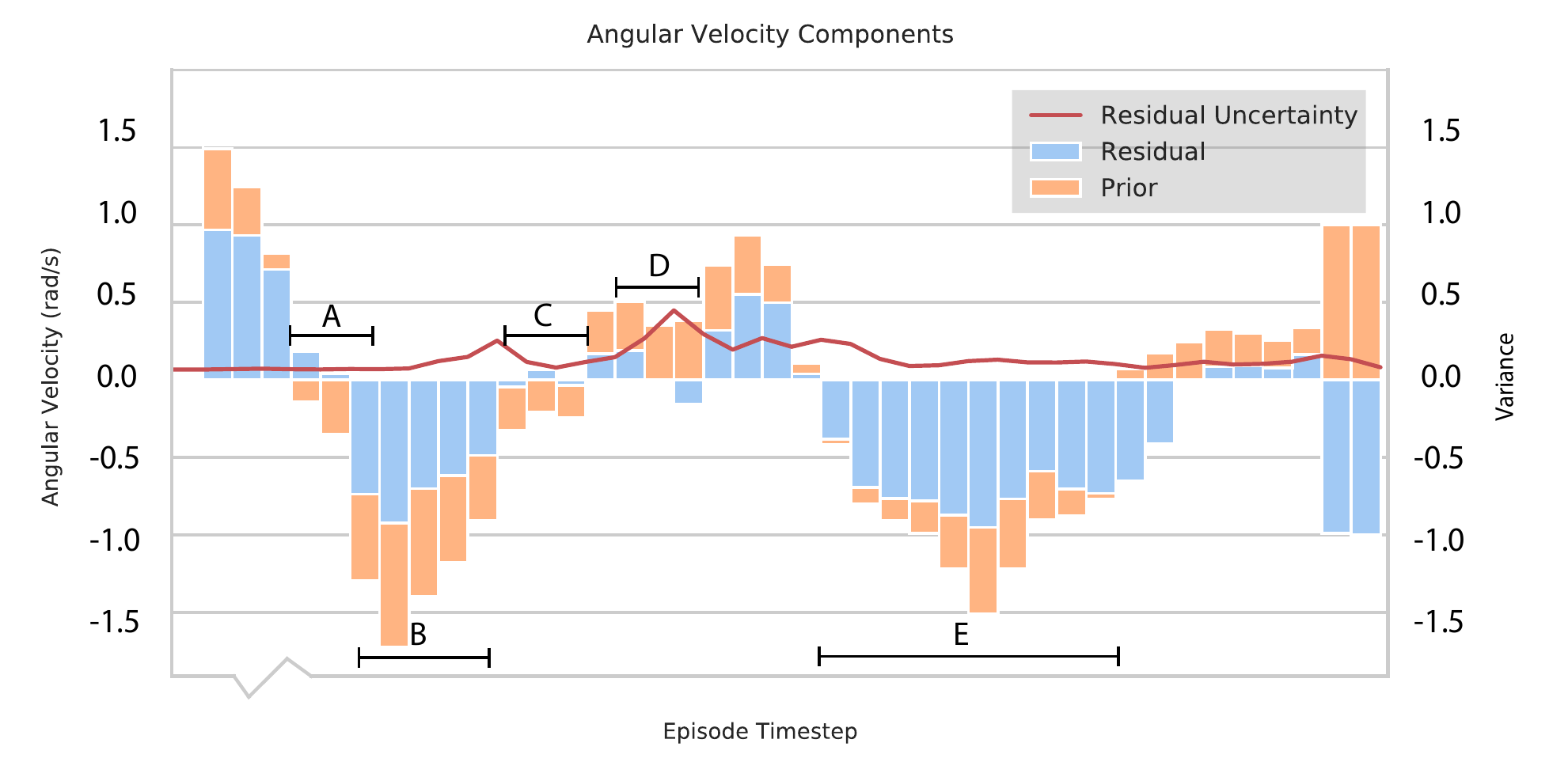}
  \caption{Stacked bar plot showing angular velocity components of a particular section of an sRRN agents' trajectory. Each coloured bar indicates the magnitude of the contribution provided by the residual policy and prior controller individually. The resultant of the two is the action effect executed in the environment. The policy uncertainty is overlayed in red. Segments of interest are marked by letters A to E.}
  \label{components}
  \vspace{-0.3cm}
\end{figure}

We conduct a study to examine the effect of the residual on the control action taken and how the uncertainty impacts a meaningful switching condition. A segment of an sRRN agent's trajectory is recorded and the corresponding action components are plotted as a stacked bar plot given in Figure \ref{components}. The length of each coloured bar indicates the magnitude of the control effect from the residual and prior individually, and the total sum of their magnitudes is the total action effect executed in the environment. We overlay the residual policy's uncertainty we extracted using dropout on the bar plots and indicate areas of interest from A to E. Segments A and C illustrate the residual policy slowing the priors' angular transition from left to right to reduce the chances of oscillations by providing a gradual opposing velocity to the prior. In cases of agreement between the two systems, the residual amplifies the prior's control effect in states it deems safe to, given its motivation to reach the goal as fast as possible. This is shown in Segments B and E. Segment D indicates a region of higher policy uncertainty and a case of prior fallback as shown by the sole orange bar in the region. We show that this ability to fallback plays a significant role when deployed on a real robot, particularly when transferred directly from simulation. We discuss this in more detail in Section \ref{sec:robot_eval}.

\subsubsection{Real Robot Evaluation}

Given the close correspondence of laser scans between simulation and the real world, we directly transfer the trained policy to a real robot for deployment after training. A PatrolBot mobile base shown in Figure \ref{front} is used, which is equipped with a 180$^{\circ}$ laser scanner. All velocity outputs were scaled to a maximum of 0.25 $m/s$ before execution on the robot. \\
The deployment environment was a typical office environment which had been previously mapped using the laser scanner. We utilise the \textit{AMCL} ROS package to localise the robot within this map and pass the relevant state information to both the prior and policy. We note here that despite the availability of a global map, the robot is only provided with global pose information and no additional information about its operational space. The environment additionally contained clutter which was unaccounted for in the mapping process. We utilise a global planner to generate target locations 3 meters apart for our local planner, enabling large traversals through the office space. \\

\begin{table}[bt]
\centering
\caption{Evaluation for real world navigation}
\label{tab:real_robot}
\resizebox{0.49\textwidth}{!}{%
\begin{tabular}{@{}lcccc@{}}
\toprule
 & \multicolumn{2}{c}{\textbf{Trajectory 1}} & \multicolumn{2}{c}{\textbf{Trajectory 2}} \\ \midrule
\multicolumn{1}{c}{\textbf{Method}} & \textbf{\begin{tabular}[c]{@{}c@{}}Distance \\ Travelled\\ (meters)\end{tabular}} & \textbf{\begin{tabular}[c]{@{}c@{}}Actuation Time\\ (seconds)\end{tabular}} & \textbf{\begin{tabular}[c]{@{}c@{}}Distance \\ Travelled\\ (meters)\end{tabular}} & \textbf{\begin{tabular}[c]{@{}c@{}}Actuation Time\\ (seconds)\end{tabular}} \\ \midrule
Prior Only & 34.6 & 256 & 32.5 & 422 \\
Move Base & Fail & Fail & 38.3 & 191 \\
End-to-end (TD3) & Fail & Fail & Fail & Fail \\
\textbf{sRRN} & \textbf{35.6} & \textbf{211} & \textbf{46.3} & \textbf{223} \\
RRN (without switching) & 45.7 & 274 & Fail & Fail \\ \bottomrule
\end{tabular}%
}
\vspace{-0.3cm}
\end{table}

\begin{figure}[t]
  \centering
  \includegraphics[width=0.49\textwidth]{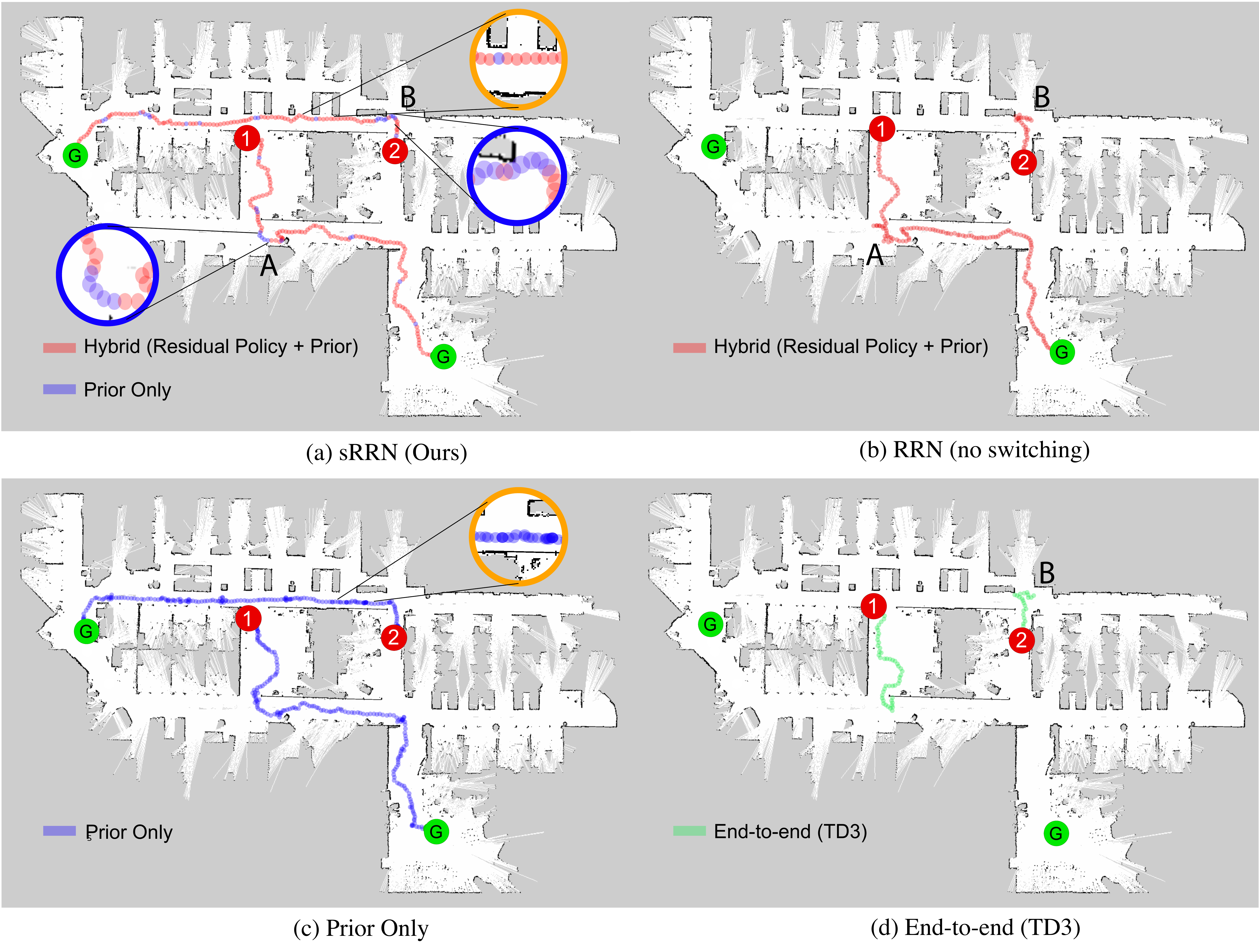}
  \caption{Trajectories taken by a real robot for two different start (red) and goal locations in a cluttered office environment with long narrow corridors. The trajectory was considered unsuccessful if a collision occurred.}
  \label{traj_robot}
  \vspace{-0.3cm}
\end{figure}

We evaluate the performance at different start and target locations indicated as Trajectory \textbf{1} and Trajectory \textbf{2} in Figure \ref{traj_robot}. Trajectory 1 exhibits a range of obstacles, tight turns and dynamic human subjects along the traversal whilst Trajectory 2 consists of a long narrow office corridor. The second scenario and dynamic subjects were not encountered in any of the training environments. For a baseline comparison, we additionally evaluate the default \textit{move-base} local planner distributed as part of the ROS navigation stack. The results are summarised in Table \ref{tab:real_robot}. We note the shorter actuation times portrayed by our sRRN system over the prior only approach, and competitive results when compared to the widely used ROS move-base planner. 

\label{sec:robot_eval}

We overlay the trajectories taken by the four systems of interest to evaluate the reasons for the success of our approach. Given the two-fold benefit sRRN can provide, we firstly compare its performance to its non-switching variant and the end-to-end learned system. \\
The sRRN trajectory, shown in Figure \ref{traj_robot} (a) is colour coded to indicate regions of prior fallback (purple) or execution of the hybrid (prior + policy) system (red). Dense regions of purple indicate higher residual policy uncertainty whereas the sparse switches seen along the trajectory are a result of the stochasiticity of the algorithm. Whilst Trajectory 1 was successful for both the RRN variants, we can see how sRRN determined a smoother trajectory at region of interest \textbf{A} as a result of prior fallback (shown by the dense regions of purple). Trajectory 2 was a difficult scenario for both the non-switching variant of RRN and the end-to-end trained policy particularly because narrow corridors were different to any of the scenarios encountered during training. Both of these approaches failed at region of interest \textbf{B}. sRRN on the other hand was capable of progressing from this region as a result of prior fallback given the residual policy's higher uncertainty in this region. We enlarge the key regions that our switching system contributed to the success of the trajectory shown within the blue outline.\\
The trajectories additionally show how learning a residual was capable of improving the efficiency of the prior. We direct the reader's attention to the enlarged regions outlined in orange shown in Figure \ref{traj_robot} (a) and (c) which illustrate the amount of time spent by the agent in a particular region. Darker regions indicate slow and inefficient oscillatory motions. It was interesting to note the reversing behaviour exhibited by RRN in the cases where it found itself in corners, or too close to obstacles. Whilst such behaviours can be hard-coded after identifying the dead-end, we show learning a residual was capable of modifying the prior's behaviours to achieve this. We provide a video to demonstrate these behaviours on our project page.

\section{Conclusions}
We show that by learning an enhancement policy on top of a competent classical controller, we can attain the performance advantages of learned systems whilst exploiting the availability of the prior to fall back on in cases of high policy uncertainty. This allows for superior performance when transferring from simulation to a real robot when compared to end-to-end based approaches. Given that we allow the residual to develop reversing behaviours, we found our robot to get stuck in particular scenarios where it blindly reversed into a wall. This is a particular limitation in our implementation which can be addressed by utilising rear-facing sensors to inform the agent. In future work we will explore alternate fusion strategies in cases where the uncertainty of the prior can be utilised. 

% LIMITATION::::i forgot to say this in the icra paper: one of the limitation of srrnn is in the ecen that in the case that the prior fails and we want to fall back in cases of policy uncertailty then our system breaks. 

% \addtolength{\textheight}{-4cm}   % This command serves to balance the column lengths

%%%%%%%%%%%%%%%%%%%%%%%%%%%%%%%%%%%%%%%%%%%%%%%%%%%%%%%%%%%%%%%%%%%%%%%%%%%%%%%%

%%%%%%%%%%%%%%%%%%%%%%%%%%%%%%%%%%%%%%%%%%%%%%%%%%%%%%%%%%%%%%%%%%%%%%%%%%%%%%%%

%%%%%%%%%%%%%%%%%%%%%%%%%%%%%%%%%%%%%%%%%%%%%%%%%%%%%%%%%%%%%%%%%%%%%%%%%%%%%%%%
\section*{ACKNOWLEDGMENTS}
The authors would like to thank Jake Bruce for the development of the simulation environment, and Robert Lee and Serena Mou for valuable and insightful discussions.

\bibliographystyle{IEEEtran}
\bibliography{references,manual}

\begin{thebibliography}{10}
\providecommand{\url}[1]{#1}
\csname url@rmstyle\endcsname
\providecommand{\newblock}{\relax}
\providecommand{\bibinfo}[2]{#2}
\providecommand\BIBentrySTDinterwordspacing{\spaceskip=0pt\relax}
\providecommand\BIBentryALTinterwordstretchfactor{4}
\providecommand\BIBentryALTinterwordspacing{\spaceskip=\fontdimen2\font plus
\BIBentryALTinterwordstretchfactor\fontdimen3\font minus
  \fontdimen4\font\relax}
\providecommand\BIBforeignlanguage[2]{{%
\expandafter\ifx\csname l@#1\endcsname\relax
\typeout{** WARNING: IEEEtran.bst: No hyphenation pattern has been}%
\typeout{** loaded for the language `#1'. Using the pattern for}%
\typeout{** the default language instead.}%
\else
\language=\csname l@#1\endcsname
\fi
#2}}

\bibitem{khatib1986real}
O.~Khatib, ``Real-time obstacle avoidance for manipulators and mobile robots,''
  in \emph{Autonomous robot vehicles}.\hskip 1em plus 0.5em minus 0.4em\relax
  Springer, 1986, pp. 396--404.

\bibitem{koren1991potential}
Y.~Koren and J.~Borenstein, ``Potential field methods and their inherent
  limitations for mobile robot navigation,'' in \emph{Proceedings. 1991 IEEE
  International Conference on Robotics and Automation}.\hskip 1em plus 0.5em
  minus 0.4em\relax IEEE, 1991, pp. 1398--1404.

\bibitem{anderson2018evaluation}
P.~Anderson, A.~Chang, D.~S. Chaplot, A.~Dosovitskiy, S.~Gupta, V.~Koltun,
  J.~Kosecka, J.~Malik, R.~Mottaghi, M.~Savva, \emph{et~al.}, ``On evaluation
  of embodied navigation agents,'' \emph{arXiv preprint arXiv:1807.06757},
  2018.

\bibitem{ross2013learning}
S.~Ross, N.~Melik-Barkhudarov, K.~S. Shankar, A.~Wendel, D.~Dey, J.~A. Bagnell,
  and M.~Hebert, ``Learning monocular reactive uav control in cluttered natural
  environments,'' in \emph{2013 IEEE international conference on robotics and
  automation}.\hskip 1em plus 0.5em minus 0.4em\relax IEEE, 2013, pp.
  1765--1772.

\bibitem{kim2015deep}
D.~K. Kim and T.~Chen, ``Deep neural network for real-time autonomous indoor
  navigation,'' \emph{arXiv preprint arXiv:1511.04668}, 2015.

\bibitem{pfeiffer2017perception}
M.~Pfeiffer, M.~Schaeuble, J.~Nieto, R.~Siegwart, and C.~Cadena, ``From
  perception to decision: A data-driven approach to end-to-end motion planning
  for autonomous ground robots,'' in \emph{2017 ieee international conference
  on robotics and automation (icra)}.\hskip 1em plus 0.5em minus 0.4em\relax
  IEEE, 2017, pp. 1527--1533.

\bibitem{zhu2017target}
Y.~Zhu, R.~Mottaghi, E.~Kolve, J.~J. Lim, A.~Gupta, L.~Fei-Fei, and A.~Farhadi,
  ``Target-driven visual navigation in indoor scenes using deep reinforcement
  learning,'' in \emph{2017 IEEE international conference on robotics and
  automation (ICRA)}.\hskip 1em plus 0.5em minus 0.4em\relax IEEE, 2017, pp.
  3357--3364.

\bibitem{tai2017virtual}
L.~Tai, G.~Paolo, and M.~Liu, ``Virtual-to-real deep reinforcement learning:
  Continuous control of mobile robots for mapless navigation,'' in \emph{2017
  IEEE/RSJ International Conference on Intelligent Robots and Systems
  (IROS)}.\hskip 1em plus 0.5em minus 0.4em\relax IEEE, 2017, pp. 31--36.

\bibitem{xie2018learning}
L.~Xie, S.~Wang, S.~Rosa, A.~Markham, and N.~Trigoni, ``Learning with training
  wheels: speeding up training with a simple controller for deep reinforcement
  learning,'' in \emph{2018 IEEE International Conference on Robotics and
  Automation (ICRA)}.\hskip 1em plus 0.5em minus 0.4em\relax IEEE, 2018, pp.
  6276--6283.

\bibitem{warren1989global}
C.~W. Warren, ``Global path planning using artificial potential fields,'' in
  \emph{Proceedings, 1989 International Conference on Robotics and
  Automation}.\hskip 1em plus 0.5em minus 0.4em\relax Ieee, 1989, pp. 316--321.

\bibitem{hwang1992potential}
Y.~K. Hwang and N.~Ahuja, ``A potential field approach to path planning,''
  \emph{IEEE Transactions on Robotics and Automation}, vol.~8, no.~1, pp.
  23--32, 1992.

\bibitem{borenstein1991vector}
J.~Borenstein and Y.~Koren, ``The vector field histogram-fast obstacle
  avoidance for mobile robots,'' \emph{IEEE transactions on robotics and
  automation}, vol.~7, no.~3, pp. 278--288, 1991.

\bibitem{antich2005extending}
J.~Antich and A.~Ortiz, ``Extending the potential fields approach to avoid
  trapping situations,'' in \emph{2005 IEEE/RSJ International Conference on
  Intelligent Robots and Systems}.\hskip 1em plus 0.5em minus 0.4em\relax IEEE,
  2005, pp. 1386--1391.

\bibitem{ng2007performance}
J.~Ng and T.~Br{\"a}unl, ``Performance comparison of bug navigation
  algorithms,'' \emph{Journal of Intelligent and Robotic Systems}, vol.~50,
  no.~1, pp. 73--84, 2007.

\bibitem{bansal2019combining}
S.~Bansal, V.~Tolani, S.~Gupta, J.~Malik, and C.~Tomlin, ``Combining optimal
  control and learning for visual navigation in novel environments,''
  \emph{arXiv preprint arXiv:1903.02531}, 2019.

\bibitem{johannink2019residual}
T.~Johannink, S.~Bahl, A.~Nair, J.~Luo, A.~Kumar, M.~Loskyll, J.~A. Ojea,
  E.~Solowjow, and S.~Levine, ``Residual reinforcement learning for robot
  control,'' in \emph{2019 International Conference on Robotics and Automation
  (ICRA)}.\hskip 1em plus 0.5em minus 0.4em\relax IEEE, 2019, pp. 6023--6029.

\bibitem{silver2018residual}
T.~Silver, K.~Allen, J.~Tenenbaum, and L.~Kaelbling, ``Residual policy
  learning,'' \emph{arXiv preprint arXiv:1812.06298}, 2018.

\bibitem{Iscen2018PoliciesMT}
A.~Iscen, K.~Caluwaerts, J.~Tan, T.~Zhang, E.~Coumans, V.~Sindhwani, and
  V.~Vanhoucke, ``Policies modulating trajectory generators,'' in \emph{CoRL},
  2018.

\bibitem{xie2017towards}
L.~Xie, S.~Wang, A.~Markham, and N.~Trigoni, ``Towards monocular vision based
  obstacle avoidance through deep reinforcement learning,'' \emph{CoRR}, 2017.

\bibitem{gal2016dropout}
Y.~Gal and Z.~Ghahramani, ``Dropout as a bayesian approximation: Representing
  model uncertainty in deep learning,'' in \emph{international conference on
  machine learning}, 2016, pp. 1050--1059.

\bibitem{kleiner2012big}
A.~Kleiner, A.~Talwalkar, P.~Sarkar, and M.~Jordan, ``The big data bootstrap,''
  \emph{arXiv preprint arXiv:1206.6415}, 2012.

\bibitem{pearce2018uncertainty}
T.~Pearce, M.~Zaki, A.~Brintrup, and A.~Neel, ``Uncertainty in neural networks:
  Bayesian ensembling,'' \emph{arXiv preprint arXiv:1810.05546}, 2018.

\bibitem{lakshminarayanan2017simple}
B.~Lakshminarayanan, A.~Pritzel, and C.~Blundell, ``Simple and scalable
  predictive uncertainty estimation using deep ensembles,'' in \emph{Advances
  in Neural Information Processing Systems}, 2017, pp. 6402--6413.

\bibitem{kahn2017uncertainty}
G.~Kahn, A.~Villaflor, V.~Pong, P.~Abbeel, and S.~Levine, ``Uncertainty-aware
  reinforcement learning for collision avoidance,'' \emph{arXiv preprint
  arXiv:1702.01182}, 2017.

\bibitem{osband2016deep}
I.~Osband, C.~Blundell, A.~Pritzel, and B.~Van~Roy, ``Deep exploration via
  bootstrapped dqn,'' in \emph{Advances in neural information processing
  systems}, 2016, pp. 4026--4034.

\bibitem{Ltjens2018SafeRL}
B.~L{\"u}tjens, M.~Everett, and J.~P. How, ``Safe reinforcement learning with
  model uncertainty estimates,'' \emph{2019 International Conference on
  Robotics and Automation (ICRA)}, pp. 8662--8668, 2018.

\bibitem{clements2019estimating}
W.~R. Clements, B.-M. Robaglia, B.~Van~Delft, R.~B. Slaoui, and S.~Toth,
  ``Estimating risk and uncertainty in deep reinforcement learning,''
  \emph{arXiv preprint arXiv:1905.09638}, 2019.

\bibitem{daftry2016introspective}
S.~Daftry, S.~Zeng, J.~A. Bagnell, and M.~Hebert, ``Introspective perception:
  Learning to predict failures in vision systems,'' in \emph{2016 IEEE/RSJ
  International Conference on Intelligent Robots and Systems (IROS)}.\hskip 1em
  plus 0.5em minus 0.4em\relax IEEE, 2016, pp. 1743--1750.

\bibitem{fujimoto2018addressing}
S.~Fujimoto, H.~van Hoof, and D.~Meger, ``Addressing function approximation
  error in actor-critic methods,'' 2018.

\bibitem{katehakis1987multi}
M.~N. Katehakis and A.~F. Veinott~Jr, ``The multi-armed bandit problem:
  decomposition and computation,'' \emph{Mathematics of Operations Research},
  vol.~12, no.~2, pp. 262--268, 1987.

\bibitem{learn_nav}
K.~Mustafa, N.~Botteghi, B.~Sirmacek, M.~Poel, and S.~Stramigioli,
  ``\BIBforeignlanguage{English}{Towards continuous control for mobile robot
  navigation: A reinforcement learning and slam based approach},''
  \emph{\BIBforeignlanguage{English}{International Archives of the
  Photogrammetry, Remote Sensing and Spatial Information Sciences}}, vol.~42,
  no. 2/W13, pp. 857--863, 6 2019.

\bibitem{choset2005principles}
H.~M. Choset, S.~Hutchinson, K.~M. Lynch, G.~Kantor, W.~Burgard, L.~E. Kavraki,
  and S.~Thrun, \emph{Principles of robot motion: theory, algorithms, and
  implementation}.\hskip 1em plus 0.5em minus 0.4em\relax MIT press, 2005.

\end{thebibliography}

\end{document}